\documentclass[preprint,12pt]{elsarticle}

\usepackage{amssymb}
\usepackage{amsmath}
\usepackage{amsthm}
\usepackage{booktabs}
\usepackage{multirow}
\usepackage{algorithm}
\usepackage{algorithmic}
\usepackage{bbm} 
\usepackage{hyperref}
\usepackage{color}
\newcommand\reczy[1]{{\color{black}#1}}
\usepackage[table]{xcolor}
\usepackage{booktabs}

\journal{Neural Networks}

\begin{document}

\begin{frontmatter}

\title{Curriculum Negative Mining For Temporal Networks}

\author[1]{Ziyue Chen}
\ead{ziyuechen@berkeley.edu}
\author[2,3]{Tongya Zheng\corref{corres}}
\ead{doujiang_zheng@163.com}
\author[3,4]{Mingli Song}
\ead{brooksong@zju.edu.cn}

\cortext[corres]{Corresponding author}

\affiliation[1]{organization={Department of Economics, University of California, Berkeley},
            city={Berkeley},
            state={California},
            country={United States}}
            
\affiliation[2]{organization={Zhejiang Provincial Engineering Research Center for Real-Time SmartTech in Urban Security Governance, School of Computer and Computing Science, Hangzhou City University},
            city={Hangzhou},
            state={Zhejiang},
            country={China}}

\affiliation[3]{organization={State Key Laboratory of Blockchain and Data Security, Zhejiang University},
            city={Hangzhou},
            state={Zhejiang},
            country={China}}

\affiliation[4]{organization={Hangzhou High-Tech Zone (Binjiang) Institute of Blockchain and Data Security},
            city={Hangzhou},
            state={Zhejiang},
            country={China}}

\begin{abstract}
Temporal networks are effective in capturing the evolving interactions of networks over time, such as social networks and e-commerce networks.
In recent years, researchers have primarily concentrated on developing specific model architectures for Temporal Graph Neural Networks (TGNNs) in order to improve the representation quality of temporal nodes and edges. However, limited attention has been given to the quality of negative samples during the training of TGNNs.
When compared with static networks, temporal networks present two specific challenges for negative sampling: \textit{positive sparsity} and \textit{positive shift}. \textit{Positive sparsity} refers to the presence of a single positive sample amidst numerous negative samples at each timestamp, while \textit{positive shift} relates to the variations in positive samples across different timestamps.
To robustly address these challenges in training TGNNs, we introduce Curriculum Negative Mining (CurNM), a model-aware curriculum learning framework that adaptively adjusts the difficulty of negative samples. Within this framework, we first establish a dynamically updated negative pool that balances random, historical, and hard negatives to address the challenges posed by \textit{positive sparsity}. Secondly, we implement a temporal-aware negative selection module that focuses on learning from the disentangled factors of recently active edges, thus accurately capturing shifting preferences. Finally, the selected negatives are combined with annealing random negatives to support stable training.
Extensive experiments on 12 datasets and 3 TGNNs demonstrate that our method outperforms baseline methods by a significant margin. Additionally, thorough ablation studies and parameter sensitivity experiments verify the usefulness and robustness of our approach.
Our code is available at \href{https://github.com/zziyue83/CurNM}{https://github.com/zziyue83/CurNM}.
\end{abstract}

\begin{keyword}
Temporal Graph Neural Networks \sep Negative sampling \sep Curriculum learning \sep Disentanglement learning
\end{keyword}

\end{frontmatter}

\section{Introduction}
Temporal networks have gained popularity in recent years due to their capacity to capture the evolving nature inherent in diverse domains, such as social networks~\cite{kn:Liu2021Enhancing, kn:Panzarasa2009uci, kn:shetty2004Enron}, recommendation systems~\cite{kn:bennett2007Netflix,kn:kumar2019jodie}, financial transactions~\cite{kn:pareja2019evolvegcn}, transport~\cite{kn:Schafer2014flights}, and political networks~\cite{kn:Fowler2006USLegis, kn:Huang2020Laplacian, kn:MacDonald2015UNTrade}. Unlike their static counterparts, these networks support additions, deletions, and changes in the features of nodes and edges over time by modeling a sequence of time-stamped events. To best leverage this temporal information, Temporal Graph Neural Networks (TGNNs)~\cite{kn:pareja2019evolvegcn,kn:kumar2019jodie,kn:rossi2020tgn,kn:sankar2019dysat,kn:manessi2020dynamic,kn:trivedi2019dyrep,kn:xu2020tgat} have been developed.

Similar to static graph models, TGNNs~\cite{kn:rossi2020tgn,kn:xu2020tgat} employ negative sampling techniques to prevent the models from being overwhelmed by a disproportionally high number of negative over positive samples and enhance the models' efficiency. The most straightforward approaches often adopted are random sampling and recent sampling~\cite{kn:rossi2020tgn,kn:xu2020tgat}. However, temporal networks introduce two unique challenges that particularly undermine these approaches: \textit{positive sparsity} and \textit{positive shift}. \textit{Positive sparsity} emerges as the inclusion of a temporal dimension significantly increases the negative-to-positive ratio by the number of timestamps, compared to static graphs. Consequently, it becomes particularly challenging to identify meaningful negative samples for training temporal networks, necessitating either the synthesis of additional positive samples or a more careful selection of negative ones.
\textit{Positive shift}, on the other hand, occurs as the relationships between nodes change over time; for instance, in recommendation systems, users' preferences are likely to evolve~\cite{kn:kumar2019jodie}. Ignoring this factor may result in training models on outdated information, particularly in datasets spanning long periods.

Nonetheless, the negative sampling problem in temporal network learning has been seldom explored. While such efforts are limited for TGNNs, there is a wealth of strategies developed for static graphs. Whether by adopting alternative distributions~\cite{kn:chen2017nncf,kn:wu2019nce,kn:zhang2013dns} or employing model-aware methods to select negatives that closely resemble positives ~\cite{kn:ding2020srns,kn:huang2021mixgcf,kn:lai2023dens,kn:zhang2019nscaching,kn:zhao2023ans}, these existing strategies seek to enhance models' performance by focusing on more informative negative samples, specifically those that closely resemble positive ones, during training. However, applying these strategies directly to temporal networks is problematic due to \textit{positive sparsity}. In particular, given the scarcity of positive samples, using negative samples that are too similar to positives can skew the classification boundary excessively towards the negatives. This can lead to confusion within the model, particularly during early epochs, and result in poor performance. Thus, a more gradual transition from easy to hard negatives is needed. Moreover, \textit{positive shift} induces a negative shift, rendering existing methods for static graphs ineffective for temporal networks. This highlights the need for components that can incorporate recent information and update dynamically over time. Till now, ENS~\cite{kn:gao2024ens} is the only negative sampling method specifically designed for temporal networks. While targeting temporal networks, ENS adopts a heuristic, global negative mining strategy instead of an adaptive, personalized approach, leading to unstable performance. Additionally, it suffers from low efficiency due to sampling from all historical edges.

To address these challenges, we propose CurNM, a curriculum learning strategy that dynamically adjusts the difficulty of negative samples based on the model's performance. This approach effectively mitigates the steep learning curves often encountered in existing methods.
Within this framework, we introduce two key components tailored to the unique characteristics of temporal networks.
First, to handle \textit{positive sparsity}, we construct a negative pool comprising a mix of random samples and historical neighbors. Hard negatives are introduced as the model demonstrates stronger learning capabilities. Starting with a smaller negative pool also ensures computational efficiency.
Second, we develop a negative selection function that incorporates temporal-aware disentanglement to capture \textit{positive shifts} and individual node preferences, further enhancing the model's granularity.

Our key contributions are summarized as follows:
\begin{itemize}
    \item We identify two crucial challenges of negative sampling in temporal networks, i.e., \textit{positive sparsity} and \textit{positive shift}.
    \item An efficient curriculum learning framework on negative sampling is tailored for temporal networks to handle these two challenges.
    \item Extensive experiments on 12 temporal network datasets using 3 TGNN models demonstrate that CurNM consistently improves link prediction performance when paired with models that capture fine-grained temporal dynamics. Its effectiveness is especially pronounced on datasets with strong temporal characteristics or significant temporal evolution.
    \item CurNM introduces minimal computational overhead, is highly portable, and can be integrated as a plug-in into any TGNN with minimal modifications, making it well-suited for practical deployment.
\end{itemize}

%--------------------------------------------------------------------------Related Work
\section{Related Work}
\subsection{Temporal Graph Learning}
Motivated by the significant progress of deep learning in Computer Vision~\cite{kn:krizhevsky2012imagenet} and Natural Language Processing~\cite{kn:mikolov2013distributed}, learning dense embeddings for nodes has become a \textit{de facto} standard paradigm in graph learning~\cite{kn:grover2016node2vec, kn:Perozzi2014DeepWalk}, unifying different graph tasks into a graph representation learning task. Subsequently, Graph Neural Networks~\cite{kn:hamilton2018GraphSAGE,kn:kipf2017GCN,kn:vel2018GAT}, which take both graph structure and node features into account, replaced skip-gram models~\cite{kn:grover2016node2vec, kn:Perozzi2014DeepWalk} for graph representations. Influenced by the advances of static graph representation learning, TGNNs have been proposed to handle the evolving dynamics of temporal networks~\cite {kn:Skarding21}, updating node embeddings as time elapses.

Existing TGNNs can be classified into discrete-time and continuous-time TGNNs based on their temporal natures~\cite{kn:kazemi2020representation}.

On the one hand, discrete-time TGNNs~\cite{kn:cong2022dynamic,kn:manessi2020dynamic,kn:pareja2019evolvegcn,kn:sankar2019dysat,kn:you2022roland,kn:zhang2023dyted} focus on the sequential dynamics of a series of graph snapshots, where the timespan of a snapshot depends on the specific graph domain.
Early approaches like EvolveGCN~\cite{kn:pareja2019evolvegcn} and DGNN~\cite{kn:manessi2020dynamic} combine recurrent models such as GRU~\cite{kn:cho2014rnn} and LSTM~\cite{kn:graves2012supervised} with GCNs, integrating temporal memory into graph learning. Transformer-based models like DGT~\cite{kn:cong2022dynamic} and DySAT~\cite{kn:sankar2019dysat} extend this by leveraging self-attention to capture both structural dependencies and temporal dynamics in evolving graphs.
More recent methods streamline dynamic graph learning using general-purpose strategies. For instance, DyTed~\cite{kn:zhang2023dyted} uses contrastive learning to disentangle temporal and stable features, and ROLAND~\cite{kn:you2022roland} introduces a meta-learning framework that transforms static GNNs into dynamic ones.

On the other hand, continuous-time TGNNs~\cite{kn:kumar2019jodie,kn:rossi2020tgn,kn:yu2023towards, kn:zhang2023tiger} focus on the sequential interactions of nodes, thereby offering more fine-grained temporal information. Various design paradigms exist for continuous-time TGNNs since temporal networks can be seen as sequential dynamics of either nodes or a graph.
JODIE~\cite {kn:kumar2019jodie} utilizes two mutually updated RNNs to update user and item embeddings, alongside a temporal attention layer that projects user embeddings after a certain time since their last interaction.
Conversely, TGN~\cite{kn:rossi2020tgn} presents a generic framework on continuous temporal networks, simulating both RNNs~\cite{kn:kumar2019jodie,kn:trivedi2019dyrep} and TGNNs~\cite{kn:xu2020tgat} with its flexible memory mechanism and GNNs.
Recent \textit{state-of-the-art} continuous-time TGNNs~\cite{kn:yu2023towards, kn:zhang2023tiger} have further explored the self-attention mechanism on temporal networks, inspired by the advancement of Transformer networks~\cite{kn:brown2020language}.
However, TGNNs typically employ random and recent negative sampling~\cite{kn:rossi2020tgn,kn:xu2020tgat}, with little attention given to the specific study of negative sampling strategies within this domain.

\subsection{Hard Negative Sampling}
Negative sampling strategies~\cite{kn:chen2017nncf, kn:ding2020srns, kn:huang2021mixgcf, kn:lai2023dens, kn:wu2019nce, kn:zhang2013dns, kn:zhang2019nscaching, kn:zhao2023ans} have been abundant in graph representation learning, such as collaborative filtering and knowledge graphs. These strategies are traditionally categorized into two types: static and model-aware negative sampling.

Static negative sampling strategies~\cite{kn:mikolov2013distributed} select negative samples based on a pre-defined probability distribution. The simplest and most common strategy is random sampling, which uniformly selects negative samples from uninteracted nodes. However, random sampling falls short in ensuring the quality of the samples. In response, alternative distributions have been proposed. For instance, NNCF~\cite{kn:chen2017nncf} and NCEPLRec~\cite{kn:wu2019nce} adopt item-popularity-based sampling distributions to preferentially select popular items as negative samples, thereby mitigating the widespread popularity bias in recommender systems~\cite{kn:chen2021bias}. Despite these advancements, all these methods fail to adequately account for distribution variations across different networks, thus leading to unsatisfactory performance.

To tackle this challenge, model-aware negative sampling strategies leverage models' specific information to oversample negatives that closely resemble positive samples, thereby tightening models' decision boundaries and enhancing accuracy. DNS~\cite{kn:zhang2013dns} pioneers this approach by dynamically choosing negative samples with the highest prediction scores. NSCaching~\cite{kn:zhang2019nscaching} builds on this by keeping a cache of hard negatives, whereas SRNS~\cite{kn:ding2020srns} introduces a variance-based criterion to reduce the risk of false negatives. Moreover, MixGCF~\cite{kn:huang2021mixgcf} synthesizes hard negatives by infusing positive information into negative samples. DENS~\cite{kn:lai2023dens} splits embeddings into relevant and irrelevant factors and selects negatives that only differ in relevant factors from the positives. ANS~\cite{kn:zhao2023ans} combines factor-aware and augmentation techniques from these strategies; it separates the hard and easy factors of negative items and synthesizes new negatives by augmenting the easy factors. While these aforementioned strategies have shown encouraging outcomes, their direct application to TGNNs is disappointing due to their oversight of temporal networks' unique characteristics, particularly \textit{positive shift} and \textit{positive sparsity}.

To the best of our knowledge, ENS~\cite{kn:gao2024ens} is the only negative sampling method specifically designed for temporal networks. ENS selects negative samples by aiming to meet a difficulty target composed of the proportion of historical neighbors and average temporal proximity. While ENS tries to tackle \textit{positive sparsity} by increasing the difficulty target by a predefined amount each epoch, its heuristic approach fails to adapt to the varying needs of different models and datasets. Moreover, although it addresses \textit{positive shift} by considering temporal proximity, this method requires evaluating all historical edges to ensure accuracy, thereby consuming substantial resources. Finally, ENS does not account for individual node preferences; it indiscriminately prioritizes all nodes that have historically appeared, without distinguishing historical neighbors, and fails to recognize that interactions are often driven by certain key factors.

\subsection{Positive Sampling}

Finally, it is worth mentioning another related line of work focused on the selection of positive rather than negative samples. For example, TASER~\cite{kn:deng2024taser} proposes a logit-based strategy that favors positive samples with higher prediction confidence, aiming to reduce noise during training. While TASER may prioritize relevant and timely interactions over outdated ones - thus partially addressing the challenge of \textit{positive shift} - it offers limited benefit in tackling the issue of \textit{positive sparsity} in temporal networks. In contrast, CurNM attempts to address both challenges by selecting informative negatives.

% --------------------------------------------------------Method
\section{Problem Statement}
\paragraph{Temporal Network}
This section formulates the problem of negative sampling in temporal networks. Let $G(V, E, T, X)$ be a temporal network where $V$ and $E$ denote the set of nodes and edges, $T$ denotes the timestamps of interactions, and $X$ denotes the node features. We denote the set of interactions by $O^{+} = \{(u,  v^{+},  t) | u,  v^{+} \in V \text{ and } t \in T\}$ where each tuple $(u,  v^{+},  t)$ records an interaction between two nodes $u,  v^{+}$ that occurs at time $t$. By convention, we call $u$ source node and $v^{+}$ destination node. These pairs form positive training samples. Conversely, we define the complement of $O^{+}$ as $O^{-} = \{(u,  v^-,  t) | u,  v^- \in V, t \in T \text{ and } (u,  v^-,  t) \notin O^{+}\}$. This set contains all node pairs $u,  v^-$ that are not connected by an edge at time $t$ and is used to generate negative samples. Our goal is to define a negative sampling strategy $f$ that selects negative samples $(u,  v^-,  t)$ from $O^{-}$ to enhance the performance of existing TGNNs.

\paragraph{Historical Neighbor}
$u,  v'$ are called neighbors at $t$ if $(u,  v',  t) \in O^{+}$. If there exists $(u,  v'',  t') \in O^{+}$ such that $t' \leq t$, $v''$ is called a historical neighbor of $u$ at $t$. In our context, we further restrict historical neighbors to nodes that are not neighbors of $u$ at the current time, i.e., $(u, v'', t) \notin O^{+}$.
For each $(u, t) \in (V, T)$, we maintain a set $H_{u, t} = \{(v_1, t_1), (v_2, t_2), ..., (v_n, t_n)\}$ that records all unique historical neighbors of $u$ at $t$ and their respective interaction timestamps. In cases where $v_i$ interacts with $u$ multiple times before $t$, we use the latest timestamp for $t_i$. Historical neighbors in our method are then drawn from this set $H_{u, t}$.

\section{Methodology}
Overall, CurNM is a curriculum learning strategy that adaptively adjusts the difficulty level of negative samples during training, based on the learning performance of TGNNs. By doing so, it effectively addresses the two unique characteristics of temporal networks and the steep difficulty curves found in existing methods.

To address the issue of \textit{positive sparsity}, CurNM first constructs a dynamic negative pool composed of historical neighbors and random negatives, with hard negatives added when the model exhibits sustained improvements.
From this pool, it then selects an increasingly difficult set of negatives as the model progresses, with the difficulty curve governed by validation performance.
The difficulty of candidate negatives is assessed using a temporally-aware disentangling technique that scores them based on their similarity to positive samples. This approach guides the model in learning which factors drive interactions and emphasizes recent node reoccurrence, allowing it to capture \textit{positive shifts} and individual-level preferences.
Finally, these selected negatives are accompanied by annealing random negatives to support robust training of TGNNs.

\begin{figure}
\includegraphics[width=\linewidth]{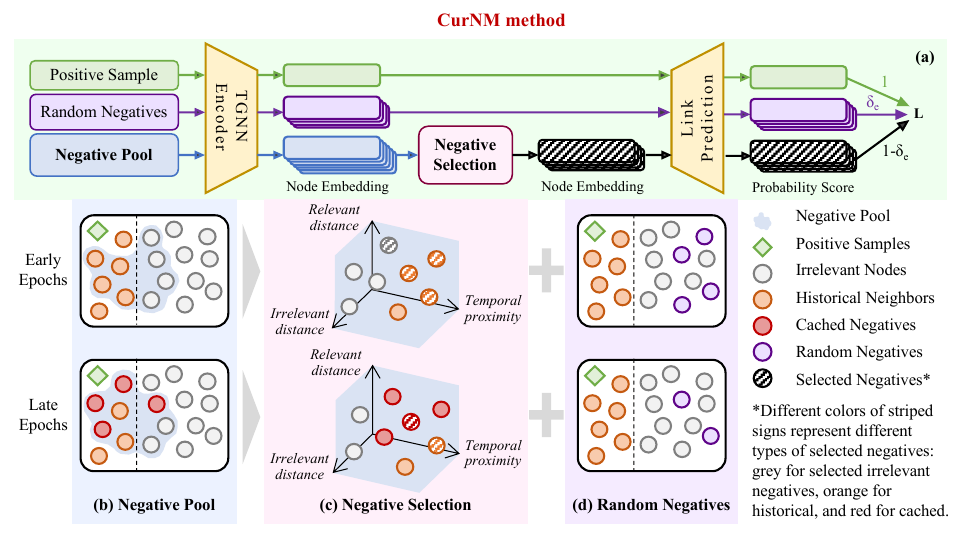}
\caption{An illustration of our CurNM method for temporal networks, which progressively increases the difficulty of sampled negatives. 
$(a)$ shows the pipeline of a single-layer TGNN, where CurNM augments basic random sampling (weighted by $\delta_e$) with selected hard negatives (weighted by $1 - \delta_e$) using a Negative Pool and a Negative Selection module. 
$(b)$ illustrates the construction of a fixed-size, dynamically updated negative pool. 
$(c)$ demonstrates the selection of negatives from this pool based on disentangled factor distances and their temporal proximity to the corresponding positive sample. 
$(d)$ presents the annealing random negatives, matched in size to the selected negatives in $(c)$, which are used together during training to ensure a smoother learning curve.
}
\label{fig:framework}
\end{figure}

This framework is illustrated in Figure~\ref{fig:framework}, and the detailed workflow is presented in Algorithm~\ref{alg:the_alg}.

\begin{algorithm}[t]
\caption{The training process with CurNM.}
\label{alg:the_alg}
\begin{algorithmic}[1]
\REQUIRE Training set \( O^+ = \{(u,  v^{+},  t)\} \), Negative set \(O^{-} = \{(u,  v^-,  t)\} \), Temporal network model \( F \).
\FOR{\(e = 1\) \TO \(E_0\)}
    \STATE Get the embeddings of nodes by \( F \).
    \STATE Sample a mini-batch \( O^+_b \) from \( O^+ \).
    \STATE Update the difficulty level $\pi_e$ by Eq.~\ref{eq:pi_update}.
    \FORALL{\( (u,  v^{+},  t) \in O^+_{\text{batch}} \)}
        \STATE Check $\mathcal{C}_e$ whether to cache training negatives by Eq.~\ref{eq:cache_epoch}.
        \IF{not $\mathcal{C}_e$}
            \STATE Get the negative pool $V_u^-$ by Eq.~\ref{eq:early_pool}.
        \ELSE
            \STATE Get the negative pool $V_u^-$ by Eq.~\ref{eq:late_pool}.
        \ENDIF
        \STATE Disentangle the relevant factors by Eq.~\ref{eq:DENS_postime} and Eq.~\ref{eq:DENS_negtime}. 
        \STATE Disentangle the irrelevant factors by Eq.~\ref{eq:DENS_irrelevant}.
        \STATE Get the scores of node pairs by Eq.~\ref{eq:sel}.
        \STATE Select informative node pairs $O^-_\text{sel}$ by Eq.~\ref{eq:selected}.
        \STATE Update embeddings by Eq.~\ref{eq:train}.
        \IF{$\mathcal{C}_e$}
            \STATE Update $V_u^{hard}$ by Eq.~\ref{eq:update_cache}.
        \ENDIF
    \ENDFOR
\ENDFOR
\end{algorithmic}
\end{algorithm}

\subsection{Dynamic Negative Pool}
First, we maintain a negative pool that dynamically balances random, historical, and hard negatives according to the model's learning performance.

To ensure efficiency, we start the entire process by sampling a negative pool $V^-_u$ of fixed size $M$ for each positive sample $(u, v^+, t)$. During the initial epochs, we sample $V^-_u$ as a mixture of historical neighbors and random negatives, a simple approach to prevent overcrowding with easy negatives.
This process can be written as
\begin{equation}
V^-_u = V^\text{hist}_u \cup V^\text{rand}_u,
\label{eq:early_pool}
\end{equation}
where $\vert V^-_u \vert=M$, and a proportion hyper-parameter $\gamma_\text{hist}$ is used to balance between $\vert V^\text{hist}_u\vert = \gamma_\text{hist} * M$ and $\vert V^\text{rand}_u\vert = (1 - \gamma_\text{hist}) * M$.

We increase the difficulty level once the model has learned the fundamentals. To track this learning progress, we define a self-adjusting proportion $\pi_e$ based on the model’s performance at epoch $e$, which decreases as performance improves and increases otherwise.\footnote{Notably, $\pi_e$ also governs the sampling proportion of negatives; its design is detailed in Section 4.2.} When $\pi_e$ drops below a predefined threshold $\tau$, the advanced stage of the negative pool is activated. This mechanism ensures that more challenging negatives are introduced only when the model is sufficiently prepared. The indicator $\mathcal{C}_e$ for this advanced stage is determined as follows:

\begin{equation}
\mathcal{C}_e = \mathbf{1}(\pi_e \leq \tau).
\label{eq:cache_epoch}
\end{equation}

In this advanced stage, we implement a hard negative cache $V^{\text{hard}}_u$ to store a subset of $V^-_u$ from the current epoch, prioritizing likely false positives and samples that exhibit limited learning progress. Technically, these are negatives with consistently high probability scores and low variance across epochs. Thus, cached negatives are sampled as
\begin{equation}
V^\text{hard}_u \subset V^-_u, \quad \text{sampled proportionally to} \quad p_{uv} - \alpha_e \cdot \mathrm{std}(P_{uv}),
\label{eq:update_cache}
\end{equation}
where $p_{uv}$ denotes the probability score of node pair $(u, v)$ at the current epoch, $P_{uv}$ records the historical scores within the last five epochs, and std($\cdot$) is set to $0$ when results are not available. $\alpha_e$ is a hyper-parameter that controls the importance of prediction standard deviation at the $e$-th epoch. To manage potential instability in early learning stages, we initially assign a modest weight to the standard deviation and gradually increase it. Specifically, we use $\alpha_e = \alpha_{\text{max}} \times \min(\frac{e}{E'}, 1)$, where $E'$ is the epoch after which the weight remains constant, and $\alpha_{\text{max}}$ is the maximum weight that standard deviations could contribute to the performance metrics.

We set $\vert V^\text{hard}_u \vert = \frac{M}{2}$, meaning that the hard negative cache accounts for half of the negative pool in the next epoch. The remaining half continues to be composed of historical neighbors and random samples, maintaining the original sampling proportion. This split ensures that the model is equally exposed to both poorly learned negatives and previously unseen ones.
Therefore, the new updating strategy for each negative pool $V^-_u$ is written as
\begin{equation}
V^-_u = V^\text{hard}_u \cup V^\text{hist}_u \cup V^\text{rand}_u,
\label{eq:late_pool}
\end{equation}
where $\frac{\vert V^\text{hist}_u \vert}{\vert V^\text{rand}_u \vert} = \frac{\gamma_\text{hist}}{1 - \gamma_\text{hist}}$ and $\vert V^\text{hist}_u \vert + \vert V^\text{rand}_u \vert = \frac{M}{2}$.

\subsection{Adaptive Negative Selection}

Given the negative pool, we select the top $\pi_e$ fraction of negative samples for model training. Section 4.2.1 explains how the sampling proportion $\pi_e$ is determined, and Section 4.2.2 describes the method used to compute the scores $s_{uv^-}$, which are used to rank the negative samples.

\subsubsection{Sampling Proportion}
As mentioned above, the sampling proportion $\pi_e$ is dynamically adjusted based on the model's performance. Specifically, to prevent the model from being overwhelmed by overly challenging examples, we start with a relatively large $\pi_e$, allowing the inclusion of both easy and hard negatives. As the model improves, this sampling proportion is gradually reduced, exposing the model to increasingly harder negatives. Conversely, if performance plateaus or degrades, $\pi_e$ can be relaxed to include easier examples again. Furthermore, we utilize a shrinkage factor $\gamma_{\text{shrink}}$ to control the adjustment steps of $\pi_e$. A high $\gamma_{\text{shrink}}$ establishes a steep difficulty curve for model learning, while a low $\gamma_{\text{shrink}}$ sets a flatter curve. This strategy is written as

\begin{equation}
    \pi_{e+1} = 
    \begin{cases}
    \pi_e - \gamma_{\text{shrink}}, & \text{If performance improves,}  \\
    \pi_e + \gamma_{\text{shrink}}, & \text{Otherwise}
    \end{cases}
    \label{eq:pi_update}
\end{equation}

In addition, instead of the traditional method of selecting a set of negative samples for each positive sample, we merge and evaluate all candidate negatives at the minibatch level. As the learning difficulties vary across source nodes, this approach allows us to focus on more informative nodes that present greater difficulty while automatically eliminating nodes that are easier to learn as $\pi_e$ shrinks. Collectively, this strategy can be formulated as:
\begin{equation}
\begin{split}
    O^-_\text{sel} = \text{arg\,top}^{N_b}_{\left\{ (u, v^-,  t) \mid (u,  v',  t) \in O^+_b, v^- \in V^-_u \right\}} s_{uv^-},
\end{split}
\label{eq:selected}
\end{equation}
where $O^-_\text{sel}$ denotes the set of selected negatives, and $s_{uv^-}$ is the scoring function used to rank the negative samples, as defined in Eq.~\ref{eq:sel}. The operator $\text{arg\,top}^{N_b}_{\left\{ (u, v^-,  t) \mid (u,  v',  t) \in O^+_b, v^- \in V^-_u \right\}}$ selects the top $N_b$ pairs $(u, v^-,  t)$ with the highest scores. Here, $N_b = \pi_e \times |b| \times M$, where $|b|$ denotes the number of edges in the minibatch $b$ and $M$ is the number of candidate negatives per edge (i.e., the size of the negative pool $V^-_u$).

\subsubsection{Negative Ranking}
We rank negative samples from the negative pool $V^-_u$ based on their proximity to the corresponding positive samples. To compute this proximity, we disentangle irrelevant factors and enhance relevant factors using time embeddings.

\paragraph{Irrelevant Factor Disentanglement with Temporal Enhancement} 
\reczy{Recognizing that temporal networks often exhibit structured and evolving interaction patterns and thus not all information is equally valuable, we aim to decouple the relevant factors driving interactions from the irrelevant ones. To achieve this, we adopt a disentanglement method inspired by DENS~\cite{kn:lai2023dens} and further integrate temporal modeling to better capture the dynamic behaviors inherent in temporal networks.}

Following DENS~\cite{kn:lai2023dens}, we first construct a positive gate for each positive sample $(u, v^+, t)$. This gate applies a linear transformation to the concatenated embeddings of the source node $h_u$ and the destination node $h_{v^+}$ to isolate the relevant factor representation $h_{v^+}^r$ of the destination node $v^+$:
\begin{equation}
h_{v^+}^r = h_{v^+} \odot \sigma(W_p [h_u \, || \, h_{v^+}] + b_p), \label{eq:DENS_pos}
\end{equation}
where $W_p$ and $b_p$ are trainable parameters shared across all positive samples, $\sigma$ denotes the sigmoid function, and $\odot$ is the element-wise product.

To better capture the unique temporal characteristics of temporal networks, we further refine the ranking process to prioritize negative samples that are either historically relevant or recently active - samples we refer to as having higher \textit{temporal proximity}. These negatives often exhibit features previously favored by the source node, making them more difficult to distinguish from true positives.

To achieve this, we incorporate Time2Vec~\cite{kn:kazemi2019time2vec}, a time embedding method designed to capture complex temporal patterns such as recurrent and periodic behaviors. Specifically, the Time2Vec encoding we use for each time point $\tau$, denoted as $\mathbf{t2v}(\tau)$, is defined as:
\[
\mathbf{t2v}(\tau)[i] = 
\begin{cases}
W^{t2v}_i \tau + b^{t2v}_i, & \text{if } i = 0, \\
\sin(W^{t2v}_i \tau + b^{t2v}_i), & \text{if } 1 \leq i \leq k,
\end{cases}
\]
where $\mathbf{t2v}(\tau)[i]$ denotes the $i$-th element of $\mathbf{t2v}(\tau)$, and $W^{t2v}_i$ and $b^{t2v}_i$ are learnable parameters. The dimension $k$ is chosen to match the dimensionality of the node embeddings.

We first use Time2Vec to encode the timestamp of each positive sample. The resulting time-enhanced factor embedding is computed as:
\begin{equation}
\hat{h}_{v^+}^r = h_{v^+}^r \odot z(\mathbf{t2v}(t_{uv^+})),
\label{eq:DENS_postime}
\end{equation}
where $t_{uv^+}$ is the timestamp of the positive sample $(u, v^+, t_{uv^+})$, and $z(\cdot)$ denotes a normalization function that prevents the temporal signal from dominating the node representation.

This factor representation $\hat{h}_{v^+}^r$, along with the embedding $h_{v^-}$ of a candidate negative $v^- \in V_u^-$, is then used to isolate the relevant factor $h_{v^-}^r$ of the negative:
\begin{equation}
h_{v^-}^r = h_{v^-} \odot \sigma(W_n [\hat{h}_{v^+}^r \, || \, h_{v^-}] + b_n),
\label{eq:DENS_neg}
\end{equation}
where $W_n$ and $b_n$ are trainable parameters shared across all candidate negatives.

Finally, we use Time2Vec to encode two timestamps that reflect \textit{temporal proximity}: $t_{uv^-}$, the timestamp of the historical interaction between $u$ and $v^-$, and $t_{v^-}$, the timestamp of the most recent occurrence of $v^-$ before the current interaction. If the negative sample is not a historical neighbor or has never appeared before, default values of 0 are passed into Time2Vec.
The final factor embedding of the negative sample is then computed as:
\begin{equation}
\hat{h}_{v^-}^r = h_{v^-}^r \odot z(\mathbf{t2v}(t_{uv^-}) \oplus \mathbf{t2v}(t_{v^-})),
\label{eq:DENS_negtime}
\end{equation}
where $\oplus$ denotes element-wise addition.

Factors other than relevant factors are denoted as irrelevant factors:
\begin{equation}
\hat{h}_{v^{+}}^{ir} = h_{v^{+}} \ominus \hat{h}_{v^{+}}^r, \quad
\hat{h}_{v^-}^{ir} = h_{v^-} \ominus \hat{h}_{v^-}^r,
\label{eq:DENS_irrelevant}
\end{equation}
where $\hat{h}_{v^{+}}^{ir}$ and $\hat{h}_{v^-}^{ir}$ denote irrelevant factors of the destination node $v^{+}$ and the candidate negative $v^-$, respectively, and $\ominus$ is element-wise subtraction.

\paragraph{Scoring Function}
As our goal is to refine the decision boundary as the model progresses, we aim to train it with negatives that increasingly resemble positive samples. We achieve this in two phases.
In the first phase, the model focuses on learning which factors are important and how they influence interactions. This involves initially aligning irrelevant factors, encouraging the model to filter out noise and identify signals.
In the second phase, we increase the difficulty by also aligning the relevant factors, pushing the model to make finer-grained distinctions between positive and negative samples.
Technically, we define the overall scoring function used to select $O^-_\text{sel}$ as:
\begin{eqnarray} 
s_{uv^-} = -\beta_e \times |s_{r,r}^{+,-}| - (2 - \beta_e) \times |s_{ir,ir}^{+,-}|, 
\label{eq:sel}
\end{eqnarray}
where $s_{g_1,g_2}^{\text{sgn}_1,\text{sgn}_2} = \text{MLP}(h_u, \hat{h}_{v^{\text{sgn}_1}}^{g_1}) - \text{MLP}(h_u, \hat{h}_{v^{\text{sgn}_2}}^{g_2})$ represents the relative score between two factor-specific representations. $\text{MLP}(\cdot,\cdot)$ computes a probability score from the given embeddings.\footnote{Whether the two inputs are added or concatenated depends on the architecture of the underlying TGNN. A common implementation uses $\texttt{ReLU}(h_u + h_v)$ followed by a linear projection.}
We refer to $|s_{r,r}^{+,-}|$ as the \textit{relevant distance}, as it measures the score difference between the relevant factors of the positive and negative samples. Similarly, $|s_{ir,ir}^{+,-}|$ is termed the \textit{irrelevant distance}.
The hyperparameter $\beta_e$ controls the balance between relevant and irrelevant factors at epoch $e$. We set $\beta_e = \min\left(\frac{e}{E''}, 1\right)$, where $E''$ is the epoch after which both factors contribute equally.

\paragraph{Supervised Disentanglement Learning}
To supervise the training of positive and negative gates, we introduce a four-component contrastive loss $L_{\text{con}}$. These components dictate that relevant factors mainly drive interactions in positive samples and the absence of interactions in negative samples, while irrelevant factors do not and may even favor negative samples. This approach is consistent with the procedure used in DENS \cite{kn:lai2023dens}, written as
\begin{equation}
L_{\text{con}} = -(s_{r,ir}^{+,+} + s_{r,r}^{+,-} + s_{ir,r}^{-,-} + s_{ir,ir}^{-,+}),
\label{eq:DENS_Conend}
\end{equation}
where the subtraction is added to perform maximization.

\subsection{Annealing Random Negatives} 
To provide the model with additional space for exploration and adaptation, especially in early epochs, each minibatch also includes $N_b$ random negatives, forming the set $O^-_\text{rand}$, in addition to the selected negative set $O^-_\text{sel}$.

A weight $\delta_e$ is applied to balance these two types. \reczy{Initially, we assign greater weight to randomly sampled negatives, allowing the model to acquire foundational knowledge through standard sampling mechanisms. As training progresses, this knowledge is then leveraged to handle selected, presumably more challenging negatives. This process is analogous to a semi-supervised learning approach~\cite{kn:li2024exploring,kn:wang2022pseudo,kn:zhao2024unlabeled}, where the model first learns from labeled data and later uses that knowledge to benefit from unlabeled data.}

To implement this annealing behavior, we define $\delta_e = \max\{\pi_e, \delta_{\text{min}}\}$. By linking $\delta_e$ to $\pi_e$, we align the influence of random negatives with the model’s performance: when the model's performance improves, $\pi_e$ decreases, and consequently, $\delta_e$ also decreases, gradually reducing the role of random negatives. $\delta_{\text{min}}$ is a hyper-parameter that controls the minimum influence random negatives should exert, ensuring both contribute to the learning process. The loss for each minibatch is computed as
\begin{equation}
    L = L_{pos} + \delta_e L_{O^-_\text{rand}} + (1-\delta_e) L_{O^-_\text{sel}} + \lambda L_\text{con} + \mu \lVert \Theta \rVert^2_2,
    \label{eq:train}
\end{equation}
where $L_{pos}$, $L_{O^-_\text{rand}}$, and $L_{O^-_\text{sel}}$ denote the mean binary cross-entropy losses for positive samples, random negatives, and selected negatives, respectively. $L_\text{con}$ is the average contrastive loss as defined in Eq.~\ref{eq:DENS_Conend}, and $\lVert \Theta \rVert^2_2$ is a regularization term.

%--------------------------------------------------------------Experiement
\section{Experiment}
\subsection{Datasets}
We evaluate the performance of our CurNM method on $12$ standardized datasets obtained from DyGLib~\cite{kn:yu2023dyglib} and TGL~\cite{kn:zhou2022tgl}. These datasets encompass a broad spectrum of domains, sizes, durations, and network features, and are intended to represent a diverse array of datasets. According to Table~\ref{tab:dataset}, which summarizes the statistics, social and interaction networks are typically sparse, while social and economic networks often exhibit high recency and repetition. These characteristics are expected due to the limited number of relationships individuals can realistically maintain and the significant impact of recent events and cyclic patterns on public attention and market dynamics. Other features, such as average time proximity, present a more diverse result across different domains. We divide these datasets into training, validation, and test sets using a chronological split with ratios of $70:15:15$ to prevent temporal data leakage during model training.

\begin{table}[H]
\caption{Statistical Overview of the Datasets. ``Average Proximity'' is the average step difference between nodes and their historical neighbors, ``Recency'' is the percentage of current interactions that replicate the last, ``Repetition'' is the proportion of repeated edges, ``$d_{\text{max}}$'' represents the maximum temporal node degree.}
\label{tab:dataset}
\centering
\resizebox{\linewidth}{!}{
\begin{tabular}{l|llrrcccllr@{}}
\toprule
\textbf{Dataset} & \textbf{Domain}      & \textbf{Bipartite} & \textbf{$|V|$}   & \textbf{$|E|$} & \textbf{Average Proximity} & \textbf{Recency} & \textbf{Repetition} & \textbf{Density}  & \textbf{$d_{\text{max}}$} & \textbf{Time Span}     \\
\midrule
\textbf{uci}     & Social      & No        & 1900  & 59835 & 58   & 34\%    & 77\%       & 5.63E-07 & 1.55E+03    & 194 days      \\
\textbf{USLegis} & Politics    & No        & 226   & 60396  & 103  & 40\%    & 72\%       & 1.98E-01 & 1.20E+03   & 11 congresses \\
\textbf{CanParl} & Politics    & No        & 735   & 74478 & 34  & 25\%    & 39\%       & 1.97E-02 & 2.38E+03   & 13 years      \\
\textbf{enron}   & Social      & No        & 185   & 125235 & 1004  & 45\%    & 98\%        & 3.23E-04 & 2.15E+04  & 4 years       \\
\textbf{WIKI}    & Social      & Yes       & 9228  & 157474 & 62  & 79\%    & 88\%       & 2.42E-08 & 1.94E+03    & 1 month       \\
\textbf{MOOC}    & Interaction & Yes        & 7047  & 411749 & 30  & 20\%    & 57\%       & 4.80E-08 & 1.95E+04   & 1 month       \\
\textbf{UNtrade} & Economics   & No        & 256   & 507497 & 890  & 83\%    & 96\%        & 4.84E-01 & 1.22E+04  & 31 years      \\
\textbf{REDDIT}  & Social      & Yes       & 10985 & 672447 & 67  & 61\%    & 88\%       & 1.67E-08 & 5.87E+04   & 1 month       \\
\textbf{myket}   & Interaction & Yes       & 17989 & 694121 & 26  & 5\%     & 21\%       & 6.18E-09 & 1.53E+04    & 6.5 months    \\
\textbf{UNvote}  & Politics    & No        & 202   & 1035742 & 1570  & 92\%    & 98\%        & 7.09E-01 & 1.59E+04 & 71 years      \\
\textbf{LASTFM}  & Interaction & Yes       & 1980  & 1293103 & 1713  & 10\%    & 88\%       & 5.14E-07 & 5.18E+04  & 1 month       \\
\textbf{Flights} & Transport   & No        & 13170 & 1927145 & 822  & 40\%    & 84\%       & 1.82E-04 & 3.47E+04   & 1 month
\\
\hline
\end{tabular}
}
\end{table}

\subsection{Baselines and Evaluation Metrics}
\paragraph{Baselines}
We compare our method against two commonly adopted negative sampling strategies: random sampling and recent sampling~\cite{kn:rossi2020tgn,kn:xu2020tgat}. We also compare with ENS~\cite{kn:gao2024ens}, the only existing method known to us that is specifically designed for temporal networks, and TASER~\cite{kn:deng2024taser}, which focuses on positive sample selection.

\begin{itemize}
    \item Random sampling selects negative samples based on a uniform distribution.
    \item Recent sampling selects the latest historical neighbors as negatives, offering a straightforward approach to addressing issues of \textit{positive sparsity} and \textit{positive shift}.
    \item ENS samples a fixed proportion of random negatives and historical neighbors while ensuring that the average temporal proximity of the selected negatives meets a predefined threshold.
    \item TASER, though primarily designed for denoising, includes a positive sample selection module. In this module, it selects positive samples with the highest predicted scores to reduce noise during training, while adding a small amount of noise to retain potentially informative but less confident positives. In our comparison, we adopt only this component.
\end{itemize}

We conduct our experiments on JODIE \cite {kn:kumar2019jodie}, TGN \cite {kn:rossi2020tgn}, and DySAT \cite {kn:sankar2019dysat}, three well-established models that exhibit superior performance on our datasets with random sampling.

\paragraph{Evaluation Metrics}
We use Average Precision (AP) as our primary evaluation metric. Specifically, we compare the AP of CurNM with the average AP of all baseline methods for each dataset-TGNN pair. We also compare all combinations of sampling methods and TGNNs for each dataset, since in practice, typically only the optimal setup is used. Additionally, given that our experiments span 12 datasets, we evaluate the overall performance of each sampling method using average rank.

These two metrics are assessed across two scenarios: transductive and inductive negative evaluation. In the transductive setting, our test includes a mix of half-random and half-historical (TRANS) negative samples. Given that our method is specifically tailored to enhance performance against hard negatives, we anticipate improved predictive accuracy for historical neighbors~\cite{kn:yu2023dyglib}. However, it is crucial to preserve the model's capability to accurately predict random negatives. In this context, TRANS provides a robust benchmark for assessing our method's ability to balance these two objectives. Conversely, the inductive test involves only nodes not seen during training and thus provides a valuable indicator of our method's generalizability to new data. This evaluation draws on code from DyGLib~\cite{kn:yu2023dyglib}.

\subsection{Experiment Setting}
For a fair comparison, we follow default implementations in TGL \cite {kn:zhou2022tgl}. In all experiments, the Adam optimizer is used with a learning rate of $0.0001$ and a mini-batch size of $200$. We perform each experiment for up to $400$ epochs and repeat it with $5$ different random seeds. All methods are executed with a 48-core CPU, 128GB of memory, and a 24G P5000 GPU.

The contrastive loss coefficient $\lambda$ and the $L_2$ regularization coefficient $\mu$ are set to ensure balanced contributions across different loss components and are used without additional tuning. The remaining hyperparameters are carefully tuned through grid search. The size of the negative pool $M$ is tested within $\{4, 8, 12, 16\}$; the proportion of historical neighbors $\gamma_{\text{hist}}$ within $\{0.3, 0.4, 0.5, 0.6, 0.7\}$; and the hard negative cache's initiation threshold $\tau$ within $\{0.3, 0.5, 0.7\}$. The stopping epoch $E'$ and the maximum weight $\alpha_{\text{max}}$ for $\alpha_e$ are explored over $\{\frac{E_0}{10}, \frac{E_0}{2}, E_0\}$ and $\{0, 0.006, 0.012, 0.024\}$ respectively, where $E_0$ denotes the maximum number of epochs. Similarly, $E''$ for $\beta_e$ and $\delta_{\text{min}}$ for $\delta_e$ are searched within $\{\frac{E_0}{10}, \frac{E_0}{2}, E_0\}$ and $\{0.3, 0.5, 0.7\}$. All chosen values are included in the code.

\subsection{Experimental Results}

As shown in Tables~\ref{tab:transductive_main_results} and~\ref{tab:inductive_main_results}, our method consistently achieves the best ranks in both historical and inductive settings, underscoring its efficacy.

\begin{table}[H]
\centering
\caption{Average Precision of link prediction across 12 datasets and three TGNNs in the transductive setting. ``AvgImp'' denotes the average percentage improvement of CurNS compared to other negative sampling methods. Cells highlighted in red indicate that CurNS achieved a statistically significant improvement, while blue cells indicate a significant deterioration. Uncolored cells reflect no significant difference. NaN indicates failed results. ``Avg Rank'' refers to the average ranks of each sampling method.
}
\label{tab:transductive_main_results}
\resizebox{\linewidth}{!}{
\scriptsize
\begin{tabular}{|l|l|ccccc|c|}
\hline
\textbf{Dataset} & \textbf{Model} & \textbf{Random} & \textbf{Recent} & \textbf{ENS} & \textbf{TASER} & \textbf{CurNM} & \textbf{AvgImp (\%)} \\
\hline
\multirow{3}{*}{\textbf{uci}} & \textbf{DySAT} & 0.763 & 0.787 & 0.763 & 0.759 & 0.767 & -0.140 \\
 & \textbf{JODIE} & 0.839 & 0.564 & 0.855 & 0.792 & 0.888 & \cellcolor{red!20}19.773 \\
 & \textbf{TGN} & 0.824 & 0.651 & 0.849 & 0.772 & 0.847 & \cellcolor{red!20}10.517 \\
\hline
\multirow{3}{*}{\textbf{USLegis}} & \textbf{DySAT} & 0.505 & 0.602 & 0.564 & 0.566 & 0.557 & 0.071 \\
 & \textbf{JODIE} & 0.492 & 0.527 & 0.501 & 0.657 & 0.494 & \cellcolor{blue!10}-8.007 \\
 & \textbf{TGN} & 0.808 & 0.797 & 0.833 & 0.699 & 0.870 & \cellcolor{red!20}11.412 \\
\hline
\multirow{3}{*}{\textbf{CanParl}} & \textbf{DySAT} & 0.586 & 0.539 & 0.597 & 0.612 & 0.578 & -0.735 \\
 & \textbf{JODIE} & 0.572 & 0.569 & 0.570 & 0.793 & 0.627 & 2.310 \\
 & \textbf{TGN} & 0.747 & 0.705 & 0.713 & 0.751 & 0.817 & \cellcolor{red!20}12.204 \\
\hline
\multirow{3}{*}{\textbf{enron}} & \textbf{DySAT} & 0.744 & 0.727 & 0.742 & 0.731 & 0.734 & -0.259 \\
 & \textbf{JODIE} & 0.800 & 0.752 & 0.801 & 0.553 & 0.788 & \cellcolor{red!20}11.006 \\
 & \textbf{TGN} & 0.771 & 0.842 & 0.824 & 0.763 & 0.877 & \cellcolor{red!20}9.855 \\
\hline
\multirow{3}{*}{\textbf{WIKI}} & \textbf{DySAT} & 0.836 & 0.844 & 0.843 & 0.836 & 0.884 & \cellcolor{red!20}5.229 \\
 & \textbf{JODIE} & 0.791 & 0.734 & 0.786 & 0.773 & 0.803 & \cellcolor{red!20}4.245 \\
 & \textbf{TGN} & 0.906 & 0.883 & 0.908 & 0.888 & 0.921 & \cellcolor{red!20}2.742 \\
\hline
\multirow{3}{*}{\textbf{MOOC}} & \textbf{DySAT} & 0.699 & 0.611 & 0.719 & 0.711 & 0.670 & \cellcolor{blue!10}-1.767 \\
 & \textbf{JODIE} & 0.670 & 0.564 & 0.677 & 0.507 & 0.657 & \cellcolor{red!20}10.242 \\
 & \textbf{TGN} & 0.764 & 0.621 & 0.772 & 0.704 & 0.782 & \cellcolor{red!20}10.208 \\
\hline
\multirow{3}{*}{\textbf{UNtrade}} & \textbf{DySAT} & 0.527 & 0.534 & 0.528 & 0.509 & 0.530 & \cellcolor{red!20}1.060 \\
 & \textbf{JODIE} & 0.609 & 0.655 & 0.611 & 0.600 & 0.616 & -0.360 \\
 & \textbf{TGN} & 0.644 & 0.660 & 0.650 & 0.596 & 0.662 & \cellcolor{red!20}4.009 \\
\hline
\multirow{3}{*}{\textbf{REDDIT}} & \textbf{DySAT} & 0.835 & 0.813 & 0.842 & 0.838 & 0.836 & \cellcolor{red!20}0.467 \\
 & \textbf{JODIE} & 0.823 & 0.781 & 0.834 & 0.839 & 0.835 & \cellcolor{red!20}1.960 \\
 & \textbf{TGN} & 0.878 & 0.848 & 0.890 & 0.881 & 0.881 & \cellcolor{red!20}0.818 \\
\hline
\multirow{3}{*}{\textbf{myket}} & \textbf{DySAT} & 0.668 & 0.738 & 0.665 & 0.671 & 0.669 & \cellcolor{blue!10}-2.171 \\
 & \textbf{JODIE} & 0.674 & 0.664 & 0.683 & 0.638 & 0.683 & 2.756 \\
 & \textbf{TGN} & 0.694 & 0.679 & 0.696 & 0.689 & 0.698 & 1.256 \\
\hline
\multirow{3}{*}{\textbf{UNvote}} & \textbf{DySAT} & 0.510 & 0.502 & 0.508 & 0.513 & 0.516 & \cellcolor{red!20}1.486 \\
 & \textbf{JODIE} & 0.684 & 0.639 & 0.696 & 0.530 & 0.695 & \cellcolor{red!20}10.307 \\
 & \textbf{TGN} & 0.681 & 0.601 & 0.693 & 0.533 & 0.700 & \cellcolor{red!20}12.901 \\
\hline
\multirow{3}{*}{\textbf{LASTFM}} & \textbf{DySAT} & 0.701 & 0.640 & 0.702 & 0.677 & 0.763 & \cellcolor{red!20}12.290 \\
 & \textbf{JODIE} & 0.536 & 0.549 & 0.552 & 0.537 & 0.570 & \cellcolor{red!20}4.862 \\
 & \textbf{TGN} & 0.742 & 0.603 & 0.740 & 0.583 & 0.716 & \cellcolor{red!20}8.778 \\
\hline
\multirow{3}{*}{\textbf{Flights}} & \textbf{DySAT} & 0.737 & 0.699 & nan & 0.728 & 0.501 & \cellcolor{blue!10}-30.482 \\
 & \textbf{JODIE} & 0.736 & 0.484 & nan & 0.715 & 0.782 & \cellcolor{red!20}25.814 \\
 & \textbf{TGN} & 0.771 & 0.573 & nan & 0.742 & 0.815 & \cellcolor{red!20}19.236 \\
\hline
\multicolumn{2}{|c|}{\textbf{Avg Rank}} & 3.11 & 3.75 & 2.36 & 3.53 & 2.03 &  \\
\hline
\end{tabular}
}
\end{table}

\subsubsection{Transductive Setting}
In the transductive setting, our method consistently outperforms existing approaches across most dataset-model pairs, achieving an overall average rank that is 0.33 points better than the second-best method, ENS, and more than one full point ahead of random, recent, and TASER negative sampling strategies. This indicates that our method excels not only at capturing historical recurrence but also maintains strong generalization on randomly sampled negatives.

\paragraph{At the model level} CurNM consistently outperforms the mean of existing methods across all 12 datasets when paired with TGN. On average, it achieves a performance improvement of 8.7\%. 

For JODIE, CurNM also performs well overall, achieving an average improvement of 7.1\%. The only statistically significant decline is observed on USLegis. In this case, since JODIE already struggles even before CurNM, reducing training difficulty - rather than increasing it - may be more effective, as evidenced by TASER's strong performance.

CurNM shows comparatively weaker performance with DySAT. As a discrete-time TGNN, DySAT relies on static, snapshot-based modeling and lacks the temporal resolution needed for CurNM to operate effectively.
This limitation is particularly evident in datasets such as myket, MOOC, and CanParl, where interactions are rarely repeated. Without an explicit memory mechanism, DySAT struggles to capture meaningful historical patterns, particularly those from further back in time, leaving CurNM with fewer temporal signals to leverage.
The case of Flights further illustrates this issue: while CurNM selects hard negatives based on temporal similarity, DySAT’s snapshot aggregation collapses temporal order. Combined with the highly structured and repetitive nature of flight schedules, this increases the likelihood that selected negatives are actually temporally adjacent positives, raising the risk of false negatives.
Nevertheless, our method still outperforms random sampling in most DySAT settings, demonstrating its robustness even under less favorable conditions.

\begin{table}[H]
\centering
\caption{Percentage improvement of the best-performing CurNM method over the best-performing baseline (across all method–TGNN combinations) for each dataset.
}
\label{tab:dataset_comparison}
\resizebox{0.6\textwidth}{!}{
\scriptsize
\begin{tabular}{|l|cc|}
\hline
\textbf{Dataset} & \textbf{Historical (\%)} & \textbf{Inductive (\%)} \\
\hline
\textbf{uci} & 3.86 & 1.49 \\
\textbf{USLegis} & 4.44 & 3.55 \\
\textbf{CanParl} & 3.03 & -1.19 \\
\textbf{enron} & 4.16 & 1.97 \\
\textbf{WIKI} & 1.43 & 1.74 \\
\textbf{MOOC} & 1.30 & -1.69 \\
\textbf{UNtrade} & 0.30 & 0.44 \\
\textbf{REDDIT} & -1.01 & 0.00 \\
\textbf{myket} & -5.42 & -13.35 \\
\textbf{UNvote} & 0.57 & -1.29 \\
\textbf{LASTFM} & 2.83 & 5.50 \\
\textbf{Flights} & 5.71 & 5.77 \\
\hline
\end{tabular}
}
\end{table}

\paragraph{At the dataset level} In Table~\ref{tab:dataset_comparison}, we compare the best-performing CurNM-model combination with the strongest baseline across all other method-model combinations to assess its practical effectiveness on each dataset. This reflects a realistic use case in which practitioners would pair CurNM with the most suitable model to achieve optimal performance.

CurNM performs particularly well on datasets characterized by frequent recurrence of historical interactions and relatively low recency, such as uci, USLegis, enron, and Flights, surpassing the best existing methods by approximately 3-6\%. This proves that our method, through the incorporation of Time2Vec-based temporal modeling, can capture complex temporal dependencies, including long-term patterns that recent sampling fails to detect. Meanwhile, in networks with high recency - typically favoring recent sampling - such as WIKI, UNtrade, and UNvote, CurNM still outperforms the strongest baselines, further confirming its ability to capture \textit{positive shift}.

CurNM also shows solid performance on user behavior-driven datasets with non-uniform and evolving temporal patterns, such as CanParl and LASTFM, achieving improvements of nearly 3\%. This is expected, as our temporally-aware disentanglement mechanism is designed to capture individual-level preference shifts over time.

Finally, in highly sparse networks like MOOC and REDDIT, CurNM also achieves performance comparable to the best baselines. The only exception is myket, where recent sampling performs particularly well with DySAT. This is because users in myket often interact with multiple apps in short bursts but rarely return to the same app, leading to low repetition but strong temporal locality. As DySAT operates on discrete snapshots without memory, recent sampling aligns better with its structure by capturing relevant negatives and avoiding the noise introduced by historical negatives, which are less meaningful in this cold-start-heavy setting.
Nevertheless, aside from this specific case, CurNM consistently matches all other methods. This demonstrates that our approach effectively circumvents the steep gradient issues common in hard negative sampling and successfully addresses \textit{positive sparsity}.

\subsubsection{Inductive Setting}
The inductive setting allows us to compare our method's generalizability against existing methods by evaluating their performances on unseen nodes. When comparing inductive to transductive settings, the average ranks of ENS and recent sampling worsen, suggesting that their heuristic strategies might not be generalizable and could lead to overfitting on seen nodes. On the contrary, the average rank of random sampling improves, indicating its robustness despite relatively weaker performance. 
CurNM also achieves a better average rank in the inductive setting, confirming that it is not only effective but also robust and highly generalizable.

\paragraph{At the model level} According to Table~\ref{tab:inductive_main_results}, CurNM achieves statistically significant average improvements when combined with TGN across all datasets, with an average gain of 6.6\%. It also shows no significant deterioration when used with JODIE, yielding an average gain of 3.9\%. DySAT remains slightly weaker, exhibiting similar patterns as in the transductive setting.

\paragraph{At the dataset level} CurNM also follows similar overall trends. The main difference is the slight performance drop in CanParl and MOOC, due to the relatively stronger performance of DySAT - an architecture less compatible with CurNM’s design. Nevertheless, even in these two cases, CurNM falls short of the best method by no more than 2\%.

\begin{table}[H]
\centering
\caption{Average Precision of link prediction across 12 datasets and three TGNNs in the inductive setting. ``AvgImp'' denotes the average percentage improvement of CurNS compared to other negative sampling methods. Cells highlighted in red indicate that CurNS achieved a statistically significant improvement, while blue cells indicate a significant deterioration. Uncolored cells reflect no significant difference.} NaN indicates failed results. ``Avg Rank'' refers to the average ranks of each sampling method.
\label{tab:inductive_main_results}
\resizebox{\linewidth}{!}{
\scriptsize
\begin{tabular}{|l|l|ccccc|c|}
\hline
\textbf{Dataset} & \textbf{Model} & \textbf{Random} & \textbf{Recent} & \textbf{ENS} & \textbf{TASER} & \textbf{CurNM} & \textbf{AvgImp (\%)} \\
\hline
\multirow{3}{*}{\textbf{uci}} & \textbf{DySAT} & 0.726 & 0.727 & 0.737 & 0.704 & 0.748 & \cellcolor{red!20}3.335 \\
 & \textbf{JODIE} & 0.636 & 0.542 & 0.648 & 0.590 & 0.652 & \cellcolor{red!20}8.403 \\
 & \textbf{TGN} & 0.677 & 0.662 & 0.700 & 0.667 & 0.705 & \cellcolor{red!20}4.351 \\
\hline
\multirow{3}{*}{\textbf{USLegis}} & \textbf{DySAT} & 0.480 & 0.584 & 0.510 & 0.503 & 0.516 & 0.041 \\
 & \textbf{JODIE} & 0.597 & 0.596 & 0.618 & 0.664 & 0.602 & -2.575 \\
 & \textbf{TGN} & 0.763 & 0.730 & 0.789 & 0.691 & 0.817 & \cellcolor{red!20}10.185 \\
\hline
\multirow{3}{*}{\textbf{CanParl}} & \textbf{DySAT} & 0.672 & 0.630 & 0.674 & 0.667 & 0.662 & 0.193 \\
 & \textbf{JODIE} & 0.533 & 0.539 & 0.530 & 0.660 & 0.564 & 0.645 \\
 & \textbf{TGN} & 0.626 & 0.624 & 0.604 & 0.610 & 0.666 & \cellcolor{red!20}8.228 \\
\hline
\multirow{3}{*}{\textbf{enron}} & \textbf{DySAT} & 0.704 & 0.698 & 0.703 & 0.693 & 0.700 & 0.065 \\
 & \textbf{JODIE} & 0.724 & 0.693 & 0.716 & 0.543 & 0.725 & \cellcolor{red!20}9.888 \\
 & \textbf{TGN} & 0.730 & 0.811 & 0.769 & 0.709 & 0.827 & \cellcolor{red!20}9.809 \\
\hline
\multirow{3}{*}{\textbf{WIKI}} & \textbf{DySAT} & 0.821 & 0.804 & 0.823 & 0.813 & 0.822 & \cellcolor{red!20}0.826 \\
 & \textbf{JODIE} & 0.701 & 0.701 & 0.712 & 0.643 & 0.698 & 1.426 \\
 & \textbf{TGN} & 0.856 & 0.837 & 0.860 & 0.848 & 0.875 & \cellcolor{red!20}2.914 \\
\hline
\multirow{3}{*}{\textbf{MOOC}} & \textbf{DySAT} & 0.684 & 0.592 & 0.687 & 0.710 & 0.653 & \cellcolor{blue!10}-1.758 \\
 & \textbf{JODIE} & 0.580 & 0.574 & 0.576 & 0.545 & 0.587 & 3.225 \\
 & \textbf{TGN} & 0.688 & 0.653 & 0.665 & 0.652 & 0.698 & \cellcolor{red!20}5.155 \\
\hline
\multirow{3}{*}{\textbf{UNtrade}} & \textbf{DySAT} & 0.538 & 0.556 & 0.539 & 0.511 & 0.545 & \cellcolor{red!20}1.674 \\
 & \textbf{JODIE} & 0.609 & 0.680 & 0.613 & 0.606 & 0.623 & -0.415 \\
 & \textbf{TGN} & 0.661 & 0.665 & 0.660 & 0.651 & 0.683 & \cellcolor{red!20}3.565 \\
\hline
\multirow{3}{*}{\textbf{REDDIT}} & \textbf{DySAT} & 0.823 & 0.814 & 0.830 & 0.828 & 0.817 & \cellcolor{blue!10}-0.813 \\
 & \textbf{JODIE} & 0.787 & 0.777 & 0.796 & 0.814 & 0.810 & \cellcolor{red!20}2.149 \\
 & \textbf{TGN} & 0.874 & 0.853 & 0.872 & 0.880 & 0.880 & \cellcolor{red!20}1.247 \\
\hline
\multirow{3}{*}{\textbf{myket}} & \textbf{DySAT} & 0.596 & 0.689 & 0.592 & 0.597 & 0.597 & \cellcolor{blue!10}-3.024 \\
 & \textbf{JODIE} & 0.530 & 0.533 & 0.550 & 0.551 & 0.531 & -1.857 \\
 & \textbf{TGN} & 0.582 & 0.539 & 0.578 & 0.577 & 0.585 & \cellcolor{red!20}2.847 \\
\hline
\multirow{3}{*}{\textbf{UNvote}} & \textbf{DySAT} & 0.511 & 0.499 & 0.509 & 0.513 & 0.519 & \cellcolor{red!20}2.137 \\
 & \textbf{JODIE} & 0.694 & 0.635 & 0.698 & 0.521 & 0.689 & \cellcolor{red!20}9.604 \\
 & \textbf{TGN} & 0.650 & 0.594 & 0.660 & 0.506 & 0.672 & \cellcolor{red!20}12.747 \\
\hline
\multirow{3}{*}{\textbf{LASTFM}} & \textbf{DySAT} & 0.709 & 0.622 & 0.709 & 0.705 & 0.748 & \cellcolor{red!20}9.325 \\
 & \textbf{JODIE} & 0.549 & 0.526 & 0.557 & 0.564 & 0.581 & \cellcolor{red!20}5.884 \\
 & \textbf{TGN} & 0.706 & 0.596 & 0.702 & 0.601 & 0.690 & \cellcolor{red!20}6.663 \\
\hline
\multirow{3}{*}{\textbf{Flights}} & \textbf{DySAT} & 0.710 & 0.689 & nan & 0.701 & 0.507 & \cellcolor{blue!10}-27.584 \\
 & \textbf{JODIE} & 0.656 & 0.605 & nan & 0.662 & 0.705 & \cellcolor{red!20}10.112 \\
 & \textbf{TGN} & 0.696 & 0.665 & nan & 0.669 & 0.751 & \cellcolor{red!20}10.988 \\
\hline
\multicolumn{2}{|c|}{\textbf{Avg Rank}} & 2.94 & 3.81 & 2.61 & 3.53 & 1.92 &  \\
\hline
\end{tabular}
}
\end{table}

\subsection{Complexity Analysis}
Theoretically, CurNM introduces a negative mining process with a time complexity of $O(|E| \times (1 + M) \times d)$ to TGNNs, which originally have a complexity of $O(L \times |E| \times n \times d)$. Here, $L$ denotes the number of layers, $|E|$ is the number of edges, and $(1 + M)$ and $n$ represent the total number of positive plus negative samples per edge with and without CurNM, respectively. Nonlinear operations are simplified to $O(d)$.

Empirically, CurNM maintains high efficiency by limiting $M$ to a relatively small negative pool size. Table~\ref{tab:eff} presents a runtime comparison between CurNM and baseline methods on the TGN model, which demonstrates the best overall performance among the three TGNNs evaluated. We first fix $M=8$ when comparing CurNM with the baselines \footnote{Based on our grid search, we found that a pool size of $M=8$ yields the best results.}, and later examine how varying $M$ affects runtime.

According to Table~\ref{tab:eff}, CurNM adds minimal computational load compared to random sampling, recent sampling, and TASER, consistently requiring about twice the time regardless of dataset size. Moreover, it is significantly more efficient than ENS on large datasets, requiring approximately one-tenth of the time on Flights, the largest dataset tested.

\begin{table}[H]
\centering
\caption{Comparison of average runtimes per epoch between CurNM and baseline methods on the TGN model. ``Ratio'' represents the ratio of CurNM runtime to each baseline method.}
\label{tab:eff}
\resizebox{\linewidth}{!}{
\scriptsize
\begin{tabular}{|l|rr|rr|rr|rr|r|}
\hline
\textbf{Dataset} & \textbf{Random} & \textbf{Ratio} & \textbf{Recent} & \textbf{Ratio} & \textbf{ENS} & \textbf{Ratio} & \textbf{TASER} & \textbf{Ratio} & \textbf{CurNM} \\
\hline
\textbf{uci} & 7.7 & 1.6 & 8.1 & 1.6 & 9.5 & 1.3 & 7.9 & 1.6 & 13.2 \\
\textbf{USLegis} & 7.6 & 2.1 & 8.2 & 1.9 & 10.7 & 1.5 & 7.9 & 2.0 & 13.0 \\
\textbf{CanParl} & 8.2 & 2.4 & 8.8 & 2.2 & 16.1 & 1.2 & 8.3 & 2.3 & 15.7 \\
\textbf{enron} & 13.0 & 2.2 & 14.0 & 2.1 & 16.5 & 1.8 & 12.8 & 2.3 & 23.8 \\
\textbf{WIKI} & 18.7 & 1.8 & 20.1 & 1.7 & 27.5 & 1.2 & 18.6 & 1.8 & 29.1 \\
\textbf{MOOC} & 42.1 & 2.1 & 47.4 & 1.9 & 169.2 & 0.5 & 42.7 & 2.1 & 73.0 \\
\textbf{UNtrade} & 54.7 & 2.3 & 63.6 & 2.0 & 148.3 & 0.8 & 58.4 & 2.1 & 108.0 \\
\textbf{REDDIT} & 77.0 & 2.1 & 86.8 & 1.9 & 310.7 & 0.5 & 78.9 & 2.0 & 144.1 \\
\textbf{myket} & 79.8 & 2.3 & 93.7 & 2.0 & 1156.6 & 0.2 & 83.3 & 2.2 & 158.9 \\
\textbf{UNvote} & 119.1 & 2.0 & 134.1 & 1.8 & 350.1 & 0.7 & 123.2 & 1.9 & 205.3 \\
\textbf{LASTFM} & 175.1 & 1.8 & 178.1 & 1.7 & 987.4 & 0.3 & 171.7 & 1.8 & 333.1 \\
\textbf{Flights} & 224.3 & 1.8 & 223.1 & 1.8 & 3236.7 & 0.1 & 219.8 & 1.9 & 446.2 \\
\hline
\end{tabular}
}
\end{table}

According to Table~\ref{tab:eff_M}, increasing $M$ has little impact on the runtime. This is primarily because the negative pool size only affects the simple prediction layer, rather than the more computationally expensive graph embedding process, and thus contributes minimally to the overall model complexity.

\begin{table}[H]
\centering
\caption{Comparison of average runtime per epoch with different negative pool sizes.}
\label{tab:eff_M}
\resizebox{0.6\linewidth}{!}{
\scriptsize
\begin{tabular}{|l|rrrr|}
\hline
\textbf{Dataset} & \textbf{M=4} & \textbf{M=8} & \textbf{M=12} & \textbf{M=16} \\
\hline
\textbf{uci} & 13.6 & 13.2 & 13.6 & 15.0 \\
\textbf{USLegis} & 14.1 & 13.0 & 13.0 & 16.2 \\
\textbf{CanParl} & 15.9 & 15.7 & 16.1 & 16.3 \\
\textbf{enron} & 26.0 & 23.8 & 24.4 & 27.3 \\
\textbf{WIKI} & 33.6 & 29.1 & 29.7 & 36.3 \\
\textbf{MOOC} & 85.2 & 73.0 & 78.2 & 85.6 \\
\textbf{UNtrade} & 111.2 & 108.0 & 111.1 & 109.2 \\
\textbf{REDDIT} & 149.9 & 144.1 & 158.0 & 179.7 \\
\textbf{myket} & 152.0 & 158.9 & 177.3 & 183.3 \\
\textbf{UNvote} & 189.3 & 205.3 & 209.1 & 224.6 \\
\textbf{LASTFM} & 289.3 & 333.1 & 335.4 & 363.1 \\
\textbf{Flights} & 427.1 & 446.2 & 461.9 & 510.9 \\
\hline
\end{tabular}
}
\end{table}

\subsection{Ablation Study}
\begin{figure}[t]
\centering
\includegraphics[width=\columnwidth]{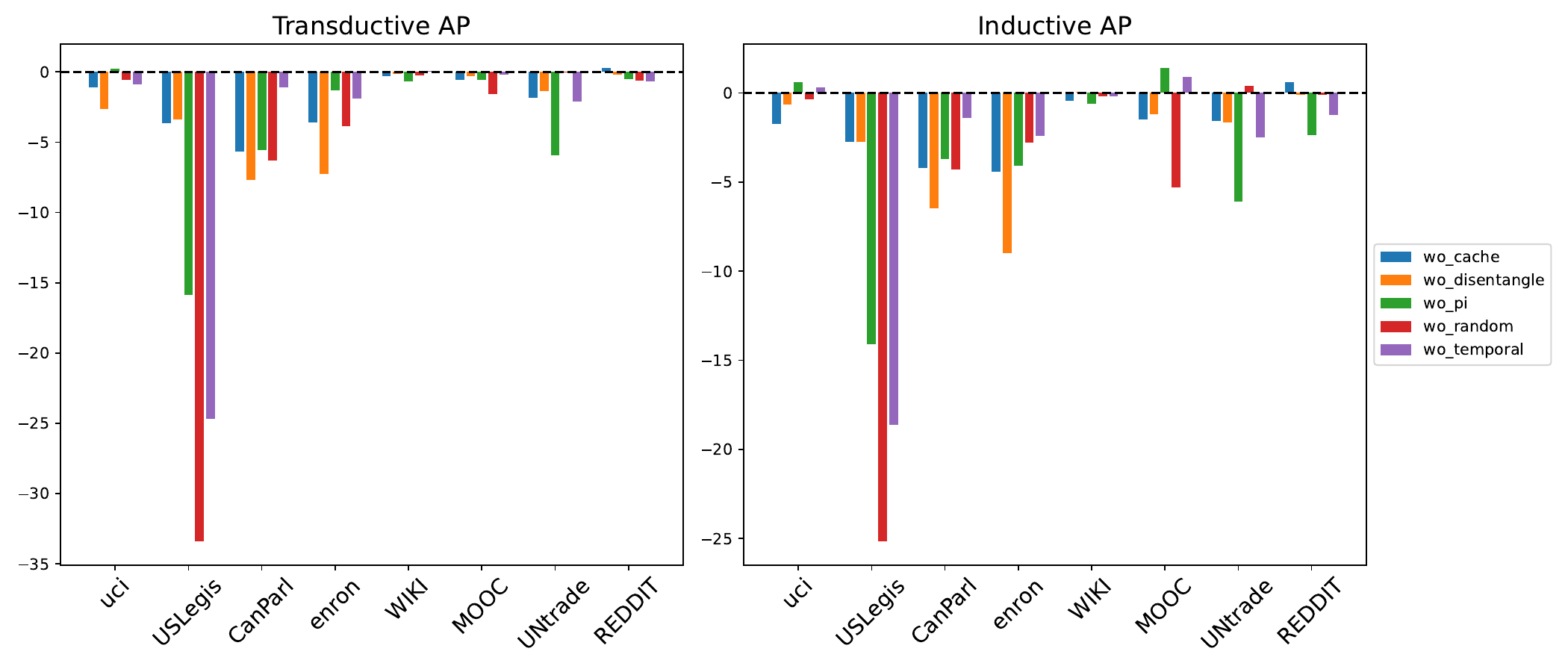}
\vspace{-1em}
\caption{Ablation studies on different components presented as percentage differences relative to the complete CurNM design.}
\label{fig:ablation}
\end{figure}

The ablation study detailed in Figure~\ref{fig:ablation} examines the effects of adaptive \underline{$\pi_e$}, annealing \underline{random} negatives, hard negative \underline{cache}, factor \underline{disentangle}ment, and \underline{temporal}-aware embeddings in our method on the TGN model. Due to space constraints, we present our study using eight datasets here.
In both transductive and inductive settings, the complete CurNM model delivers the most robust performance across the datasets.

The most significant decline in performance is observed when random negatives are removed, validating our claim that a steep difficulty curve can confuse the model.
In datasets like USLegis, CanParl, and enron, where our model achieves strong results, the absence of temporal-aware embeddings and adaptive $\pi_e$ also leads to substantial performance drops. This underscores their critical role in addressing the networks' \textit{positive sparsity} and \textit{positive shift}.
Although the impact of removing disentanglement and hard negative cache is less pronounced, their absence can still result in up to a 9\% performance decline, suggesting their importance in enhancing the granularity of learning within the model.

\subsection{Hyper-Parameter Sensitivity}
\begin{figure}[t]
\centering
\includegraphics[width=\columnwidth]{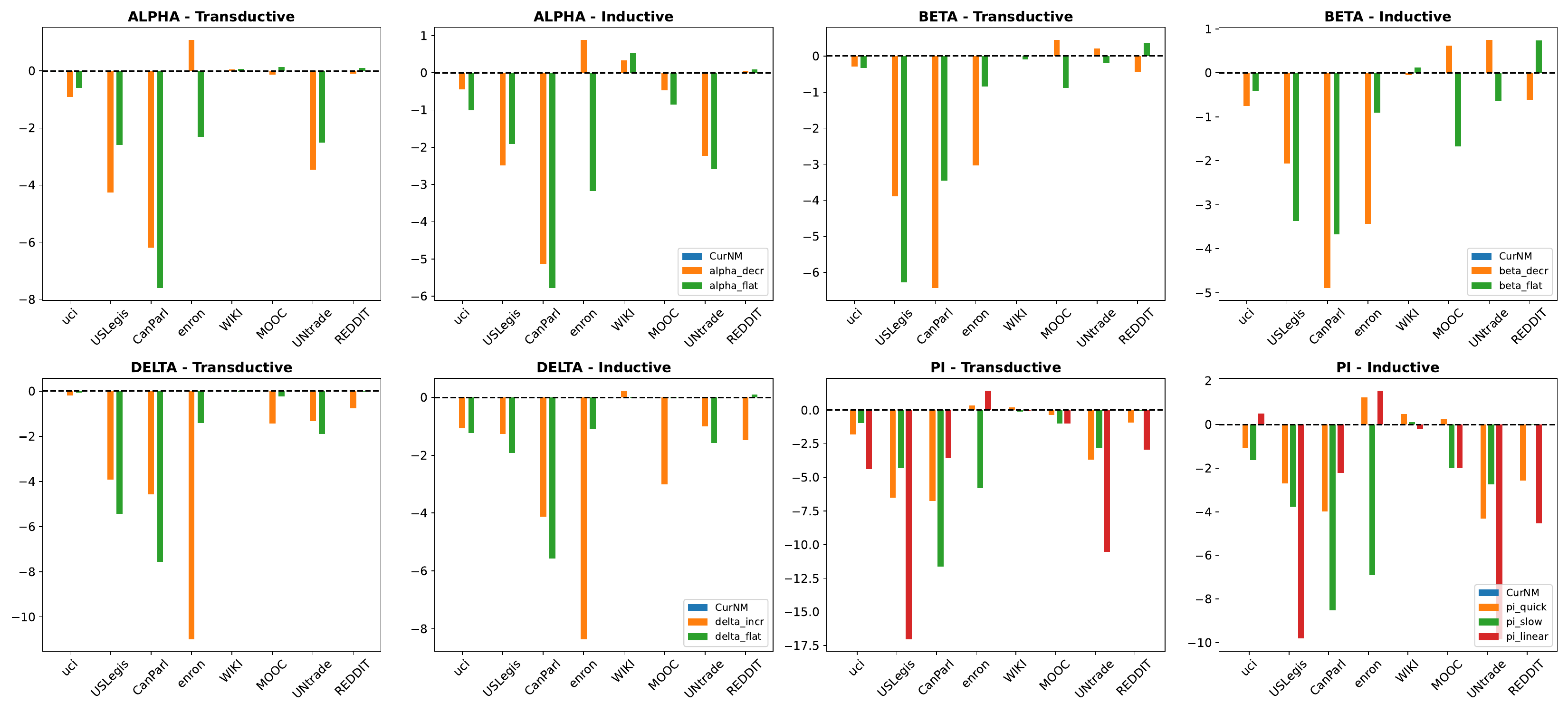}
\vspace{-1em}
\caption{Parameter sensitivity studies on $\alpha_e$, $\beta_e$, $\delta_e$, and $\pi_e$ presented as percentage differences relative to the complete CurNM design.}
\label{fig:sensitivity}
\end{figure}

This section tests our hypotheses underlying the key parameters that govern the curriculum learning process.
\paragraph{Impact of Strategies for $\alpha_e, \beta_e$, and $\delta_e$}
We investigate our assumptions about the structures of $\alpha_e, \beta_e$, and $\delta_e$ by testing an increasing strategy (e.g., $\alpha_e = \min(\frac{e}{E'}, 1)$), a decreasing strategy (e.g., $\alpha_e = \max(0, 1-\frac{e}{E'})$), and a constant strategy (e.g., $\alpha_e = 0.5$). Our findings in Figure~\ref{fig:sensitivity} confirm that, as expected, our proposed approach yields the most robust results. This highlights the effectiveness of our dynamic strategies in guiding the model gradually toward more relevant and difficult nodes, leading to a smoother and more efficient learning curve.

\paragraph{Impact of Strategies for $\pi_e$ and $\gamma_{\text{shrink}}$}
Under the framework where $\pi_e$ should shrink as learning progresses, we explore two approaches: a linear strategy and an adaptive strategy. The linear strategy consistently reduces the $\pi_e$ by a set amount each epoch, whereas the adaptive strategy only does so when there's an improvement in the selected performance metric. For our baseline, both methods start with $\pi_e = 100\%$, with a shrinkage factor $\gamma_{\text{shrink}}=3\%$, and the $\pi_e$ can decrease to a minimum of $10\%$. The choice of $\gamma_{\text{shrink}}=3\%$ is based on that models generally stabilize after approximately $30$ epochs, and having a larger $\pi_e$ before this phase to capture more information would be beneficial. The adaptive strategy outperforms the linear one. We then experiment with a slower ($\gamma_{\text{shrink}}=2\%$) and a faster shrinkage factor ($\gamma_{\text{shrink}}=5\%$). Figure~\ref{fig:sensitivity} shows that $\gamma_{\text{shrink}}=3\%$ yields the best results.

\section{Conclusion}
In this paper, we identify two unique challenges of negative sampling in temporal networks – \textit{positive sparsity} and \textit{positive shift}. Our proposed CurNM method employs a dynamically updated negative pool and a curriculum learning strategy to adaptively adjust the difficulty level of negative samples during training, thus addressing the common challenge of \textit{positive sparsity}. Furthermore, we implement a temporal-aware negative selection function to effectively capture \textit{positive shift}. Our method's efficacy is demonstrated through extensive experiments against 3 algorithms and 12 datasets.
Although CurNM demonstrates superior performance, it requires tuning several hyperparameters to optimize its performance. In the future, we aim to simplify these designs to reduce the need for extensive parameter tuning.

\section*{Declaration of interests}
The authors declare that they have no known competing financial interests or personal relationships that could have appeared to influence the work reported in this paper.

\section*{Acknowledgements}
\reczy{This work was supported by the Zhejiang Provincial Natural Science Foundation of China [Grant No. LMS25F020012], the Joint Funds of the Zhejiang Provincial Natural Science Foundation of China [Grant No. LHZSD24F020001], National Natural Science Foundation of China [Grant No. 62476238], Zhejiang Province High-Level Talents Special Support Program ``Leading Talent of Technological Innovation of Ten-Thousands Talents Program" [Grant No. 2022R52046], and the advanced computing resources provided by the Supercomputing Center of Hangzhou City University.}

\bibliographystyle{apalike}

\end{document}